\documentclass[10pt,twocolumn,letterpaper]{article}

\usepackage{cvpr}
\usepackage{times}
\usepackage{epsfig}
\usepackage{graphicx}
\usepackage{amsmath}
\usepackage{amssymb}

\usepackage{url}
\usepackage{graphicx}
\usepackage{amsmath}
\usepackage{accents}
\usepackage{statex}
\usepackage{amssymb}
\usepackage{mathrsfs}
\usepackage{cases}
\usepackage{bm}
\usepackage{color} 
\usepackage{algorithm}
\usepackage{algorithmic}
\usepackage{multirow}
\usepackage{subfigure}
\usepackage{wrapfig}

\newcommand{\prox}{\textit{prox}}

\DeclareMathOperator*{\argmax}{argmax}
\DeclareMathOperator*{\sign}{sign}

\DeclareMathOperator*{\tr}{tr}

\newcommand{\mh}{\hat M}

\newcommand{\mb}{\bar M}
\newcommand{\by}{\bar Y}
\newcommand{\bi}{\mathbb I}

\ifx\proof\undefined
\newenvironment{proof}{\par\noindent{\bf Proof\ }}{\hfill\BlackBox\\[2mm]}
\fi

\ifx\theorem\undefined
\newtheorem{theorem}{Theorem}
\fi
\ifx\assumption\undefined
\newtheorem{assumption}[theorem]{Assumption}
\fi
\ifx\condition\undefined
\newtheorem{condition}{Condition}
\fi
\ifx\example\undefined
\newtheorem{example}{Example}
\fi
\ifx\property\undefined
\newtheorem{property}{Property}
\fi
\ifx\lemma\undefined
\newtheorem{lemma}[theorem]{Lemma}
\fi
\ifx\proposition\undefined
\newtheorem{proposition}[theorem]{Proposition}
\fi
\ifx\remark\undefined
\newtheorem{remark}{Remark}
\fi
\ifx\corollary\undefined
\newtheorem{corollary}[theorem]{Corollary}
\fi
\ifx\definition\undefined
\newtheorem{definition}[theorem]{Definition}
\fi
\ifx\conjecture\undefined
\newtheorem{conjecture}[theorem]{Conjecture}
\fi
\ifx\fact\undefined
\newtheorem{fact}[theorem]{Fact}
\fi
\ifx\claim\undefined
\newtheorem{claim}[theorem]{Claim}
\fi

\usepackage[pagebackref=true,breaklinks=true,letterpaper=true,colorlinks,bookmarks=false]{hyperref}

\cvprfinalcopy 

\def\cvprPaperID{1367} 

\ifcvprfinal\fi
\begin{document}

\title{Monocular 3D Pose Recovery via Nonconvex Sparsity  with Theoretical Analysis}

\author{Jianqiao Wangni\\
University of Pennsylvania\\
{\tt\small wnjq@seas.upenn.edu}
\and
Dahua Lin\\
CUHK \\
{\tt\small dhlin@ie.cuhk.edu.hk}
\and
Ji Liu\\
University of Rochester\\
{\tt\small ji.liu.uwisc@gmail.com}
\and
Kostas Daniilidis\\
University of Pennsylvania\\
{\tt\small kostas@seas.upenn.edu}
\and
Jianbo Shi\\
University of Pennsylvania\\
{\tt\small jshi@seas.upenn.edu}
}

\maketitle

\begin{abstract}

For recovering 3D object poses from 2D images, a prevalent method is to pre-train an over-complete dictionary $\mathcal D=\{B_i\}_i^D$ of 3D basis poses. During testing, the detected 2D pose  $Y$ is matched to dictionary  by $Y \approx  \sum_i M_i B_i$ where $\{M_i\}_i^D=\{c_i \Pi R_i\}$, by estimating the rotation $R_i$, projection $\Pi$ and sparse combination coefficients $c \in \mathbb R_{+}^D$. In this paper, we propose non-convex regularization $H(c)$ to learn coefficients $c$, including novel leaky capped $\ell_1$-norm regularization (LCNR),
\begin{align*}
H(c)=\alpha \sum_{i } \min(|c_i|,\tau)+ \beta \sum_{i } \max(| c_i|,\tau), 
\end{align*}
where $0\leq \beta \leq \alpha$ and $0<\tau$ is a certain  threshold, so the invalid components smaller than $\tau$ are composed with larger regularization and other valid components with smaller regularization. We propose a multi-stage optimizer with convex relaxation and ADMM. We prove that the estimation error $\mathcal L(l)$  decays w.r.t. the stages $l$, 
\begin{align*}
Pr\left(\mathcal L(l) < \rho^{l-1} \mathcal L(0) + \delta \right) \geq  1- \epsilon,
\end{align*}
where $0< \rho <1, 0<\delta, 0<\epsilon \ll 1$. Experiments on large 3D human datasets like H36M are conducted to support our improvement upon previous approaches.  To the best of our knowledge, this is the first theoretical analysis in this line of research, to understand how the recovery error is affected by fundamental factors, e.g. dictionary size, observation noises, optimization times. We characterize the trade-off between speed and accuracy towards real-time inference in applications.
\end{abstract} 

\section{Introduction}

The computer vision community views object detection and recognition as an important task. Additionally, studying the problem in 3D geometry \cite{hartley2003multiple} has been popular since it enriches the information obtained from 2D images. 
For relatively rigid objects like human, hands, and cars, their 3D information is strongly reflected by the 3D poses, which compose of the 3D locations of important landmarks, e.g. elbow, knees and shoulders (for human). 
The recent progress in deep learning based algorithms to 3D object pose recovery could be roughly divided into two lines: 1), a two-stage pipeline that first recovers the 2D poses using specifically designed deep networks for special objects like human \cite{newell2016stacked}\cite{pishchulin2016deepcut} \cite{wei2016convolutional}\cite{chu2016structured}, and then estimate the 3D poses by solving a geometric inference problem that the 2D pose is captured through a projection, rotation, and translation of the object in the 3D world \cite{park20163d}; \cite{zhou2016sparseness} encourages the consistency of the combination coefficients and rotation matrices between two consecutive frames in a video;  Bogo et al.\cite{bogo2016keep} uses 2D detectors for preprocessing and generate a 3D mesh by the given pose, then optimize the objective function w.r.t. the error of projected 2D pose from the real pose; 2), an end-to-end pipeline that uses a deep network to regress the 3D landmarks from the images. For examples, Wu et al.\cite{wu2016single}Zhou et al.\cite{zhou2016deep} directly inference the camera matrices or kinematic model parameters using a deep network; Pavlakos et al.\cite{pavlakos2017coarse} \cite{pavlakos2018ordinal} predict the 3D landmarks by considering a ranking loss of by the ordinal relationship built upon estimated depth. In prior to these methods, detecting general objects in the wild could be accomplished by region-based CNN \cite{girshick2014rich} \cite{girshick2015fast}\cite{ren2017faster}.

In the two-stage algorithms, the part of recovering 3D poses from 2D poses actually provides a more elegant and challenging mathematical problem, since both projection parameters and 3D poses are unknown, the problem is actually ill-posed. Active pose model (ASM) \cite{cootes1995active} is a pioneering work in this area, it represents the pose as a dense combination of learned basis poses from a small dictionary, therefore the workload of searching over all nonrigid deformation is greatly reduced. 
Additionally, a sparse combination of poses from an over-complete dictionary is more powerful dealing with complex variations such as human poses. This idea resembles compressive sensing \cite{candes2006robust} that uses much less bits to transmit more information. 
Even we do have strong tools to optimize a sparse model\cite{buhlmann2011statistics}, we should note that our problem is still hard to solve, since the joint estimation of pose and viewpoint is a non-convex problem, and the orthogonality constraint on the camera rotation matrices makes it more complicated.

In this work, we will concentrate on using the two-stage pipeline with well-trained 2D detectors and a separate 3D pose inference stage. This takes advantage of massive datasets like COCO \cite{lin2014microsoft} and MPII \cite{andriluka20142d} that are only provided with 2D landmarks annotations, which are  much easier to obtain than 3D landmarks, and adapt to a wild environment like random streets. Since 3D pose datasets like H36M \cite{h36m_pami}, although available for academic usage, should cost expensive efforts to capture and label. Plus, a separate geometrical inference procedure generates a physically and mathematically reasonable result that generalize well, and help us to diagnose problems if being wrongly applied to different data sources. By incorporating this framework, previous research in 2D pose analysis with diverse motivations can be applied here, by considering the anatomical human structure like the length of limps \cite{ramakrishna2012reconstructing} \cite{wang2014robust}, joint-angle constraints \cite{akhter2015pose}, 
and temporal information \cite{zhou2014spatio}\cite{cao2017realtime}.

Our framework consists of a standard 2D landmark detector and incorporates an over-complete dictionary, and try to learn better and faster sparse coefficients during testing, which is important for real-time applications. 
Previous research mostly did not optimize the inference speed problem explicitly, nor did they provide theoretical analysis on convergence and recovery error. 
We consider the problem based on the sparsity inducing regularization, like $\ell_1$ norm. From an optimization perspective, in the standard proximal algorithm, a larger regularization will leverage the threshold of effective weights, forcing more small weights to be zero, but it also causes the effective weights deviating from ground-truth. Following this thought, we propose to use non-convex regularization to recover 3D poses. We also present a novel \emph{Leaky Capped $\ell_1$ Norm Regularizer}. We derive a multi-stage algorithm suitable for both convex and non-convex loss functions. We also provide a theoretical analysis to compensate existing empirical studies. 


\section{Monocular 3D pose Recovery}

\begin{figure}[!htbp]
	\subfigure	{\includegraphics[width=0.5\textwidth]{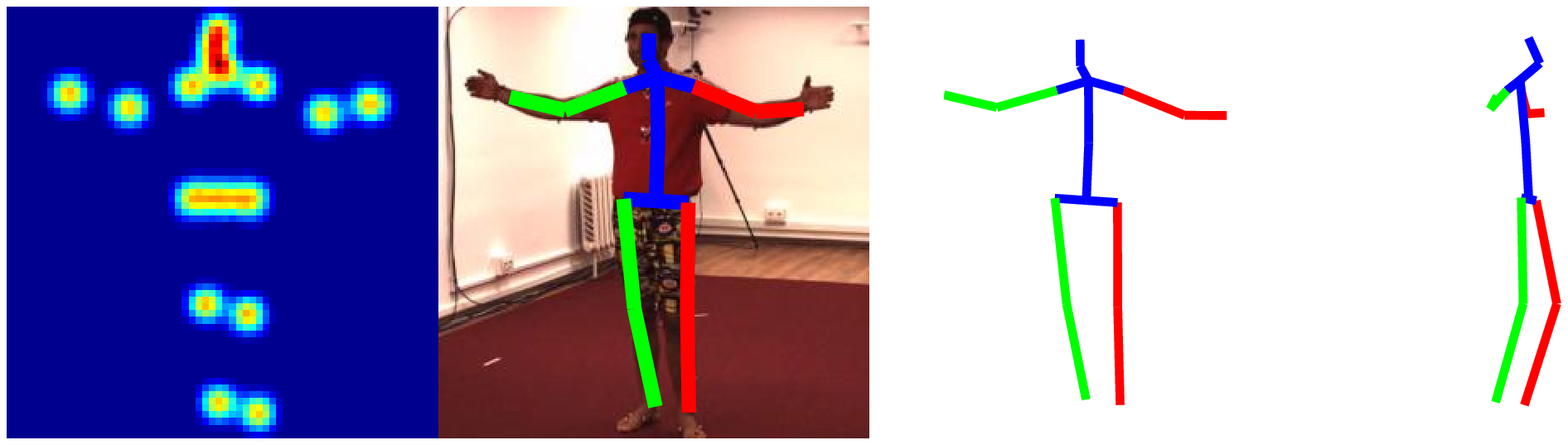}
	}
	\subfigure{
		\includegraphics[width=0.5\textwidth]{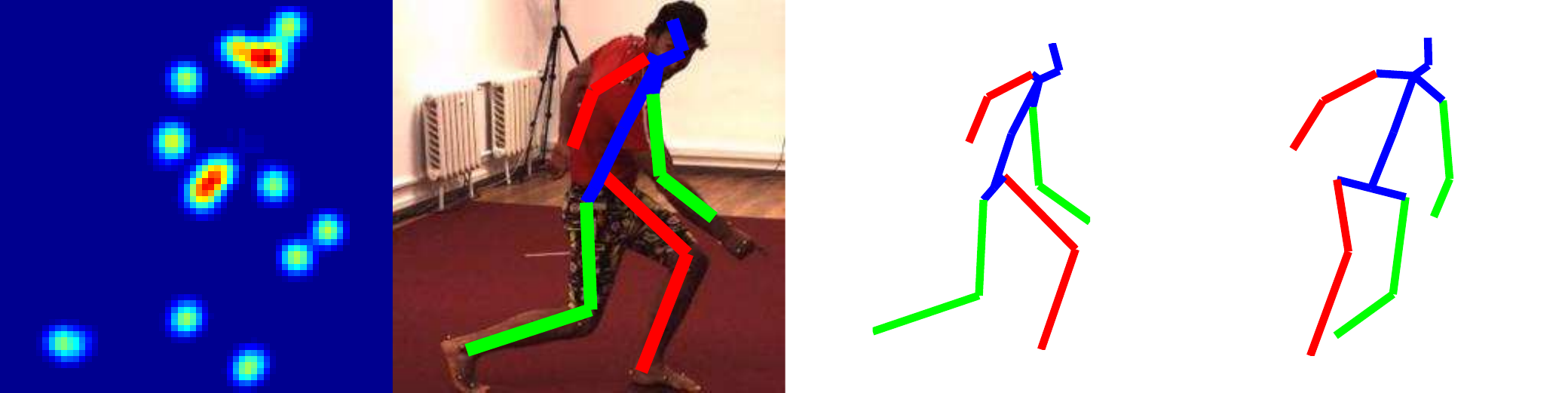}}
	\subfigure{
		\includegraphics[width=0.5\textwidth]{untitled4.eps}}
	\caption{ Recoverd results on Human36M. (From left to right: heatmaps, 2D pose, recoverd 3D pose, 3D pose in a new view. )}
	\vspace{-10pt}
	\label{fig:h36mresult}
\end{figure}

Now we give a brief introduction on the representative method based on sparse representation, which is effective in real-world applications, often demonstrating great generalization performance and higher robustness against huge variations. The 3D pose, i.e. the 3D locations of $p$ landmarks are stacked as $S \in \mathbb R^{3 \times p}$, and its corresponding 2D pose is $Y \in \mathbb R^{2 \times p}$.  The Active Shape Models (ASM) \cite{cootes1995active} proposed training a group of pose bases $\{B_i \}_{i=1}^D$ from data. 
Denoting $\mathcal D=\{1,\cdots,D\}$ as a set of subscripts and $c$ as weights of each basis,
\begin{eqnarray}
S=\sum_{i \in \mathcal D} c_i B_i ,\quad  c \in \mathbb R^{D}, B_i \in \mathbb R^{3 \times p}.
\end{eqnarray}
the 3D-2D projection is characterized as 
\begin{eqnarray}
Y= \Pi S,\quad \Pi= \begin{pmatrix} \omega &0 &0 \\0 &\omega &0 \end{pmatrix}, 
\end{eqnarray}
where we denote the projection matrix as $\Pi$, and $\omega$ is a parameter depending on physical factors like focal length and view depth. In the test phase, the 2D pose $Y$ is annotated by regular visual detectors, since the pose bases $B$ are most likely predefined in a different camera setting other than the test setting, the unknown factors including combination weights $c$, a relative rotation parameter $R$ and a translation parameter $T$ should all be inferred from
\begin{eqnarray}
Y=\Pi(R \sum_{i \in \mathcal D} c_i B_i +T), \where R \in SO(3), T \in \mathbb R^{3}
\end{eqnarray}
and $I_3 \in \mathbb R^{3 \times 3}$ is an identity matrix and 
\begin{align*}
SO(3)=\{ R \in \mathbb R^{3 \times 3}\mid R^\top R=I_3,\det(R)=1\}.
\end{align*}
In addition, \cite{zhou2016sparse} proposed distributing individual rotation matrix $R_i$ to each basis,  then the 2D poses are represented as $Y= \sum_{i \in \mathcal D} c_i R_i B_i,\quad R_i \in \mathbb R^{3 \times 3}$. 
They substitute the bilinear term composed of $\Pi$ and $R$ by uniform variables $\{M_i \in \mathbb R^{2 \times 3} \}_{i \in \mathcal D}$, that $M_i=c_i \Pi R_i$ which implicitly take rotation and projection factors into account. Denoting $M$ as a 3 dimensional tensor stacking $\{M_i\}_{i=1}^D$, we rewrite the objective as
\begin{eqnarray}\label{eq:linearf}
\min_{R_i \in \Omega(c_i),c} F(M,c) =\frac{1}{2}|| Y-  \sum_{i \in \mathcal D} M_i B_i ||_F^2 +H(c),\\
\where \Omega(c_i) =\{ M_i \in \mathbb R^{2 \times 3}| M_i^\top M_i=c_i^2 I_2 \} \nonumber, 
\end{eqnarray}
and $H(c)$ is the regularization and $I_2 \in \mathbb R^{2 \times 2}$ is an identity matrix. Here the translation $T$ is eliminated by centralizing the data. In our model of 3D pose recovery, we adopt the linear formulation as Eq.(\ref{eq:linearf}) since it is the best baseline up until now and has some attractive properties.

The model described above assumes that a high-precision 2D pose is given by the detector, in some case, this model could be improved by considering the uncertainty of 2D poses, or using the video context for such images. For example, give a sequence of images as $I=\{I_1,I_2,\cdots,I_n\}$, Monocap \cite{zhou2018monocap} considers optimizing the likelihood w.r.t the parameters $\theta =\{M,c\}$ and variance parameter $\nu$,
\begin{align}
Pr(I,Y|\theta)=Pr(I|Y)Pr(Y|\theta),
\end{align}
where $Pr(I|Y)$ can be modeled by a normalized CNN heatmap extracted from given images, and 
\begin{align}
Pr(Y|\theta)=\exp( -\frac{1}{2\nu^2} ||Y-  \sum_{i \in \mathcal D} M_i B_i ||^2 ).
\end{align}
The consistence of human poses over the sequence can be modeled by
\begin{align}
\mathcal L_{t}(\theta)=H(c)+\varphi ||\nabla_t c||_F^2 + \varGamma ||\nabla_t R||_F^2,
\end{align}
where $\nabla_t$ means taking derivatives w.r.t. the time variable of two consecutive frames in a sequence, and $\varphi,\varGamma$ are constants. Combining the terms above together, the parameter is optimized by
\begin{align}\label{em}
\argmax_{\theta} \ln Pr(Y|\theta)- \mathcal L_{t}(\theta).
\end{align}

\section{Leaky Capped $\ell_1$ Norm Regularizer }

\begin{figure}[!htbp]
  \vspace{-10pt}
  \begin{center}
    \includegraphics[width=0.5\textwidth]{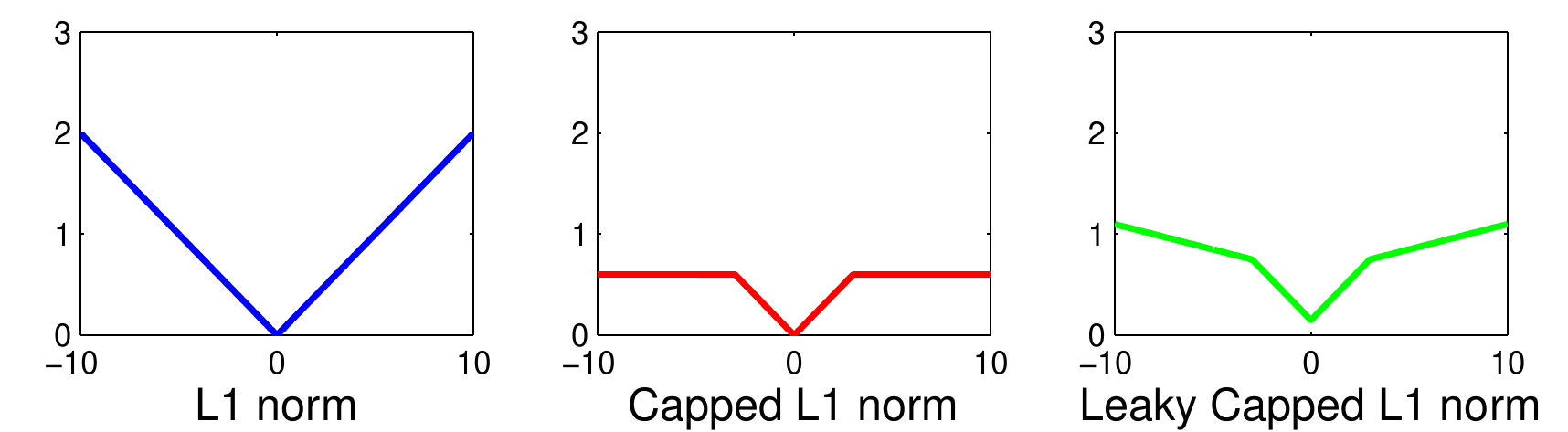}
  \end{center}
  \vspace{-5pt}
 \caption{Geometric view of three kinds of norms. }\label{fig:norm}
 \vspace{-5pt}
\end{figure}

Now we proceed to study the regularization of the aforementioned model. A popular choice for $H$ is the $\ell_1$ norm, due to its convenience for optimization and as a good surrogate of the $\ell_0$ norm. In this work, we aim to move beyond the limitations of existing literatures and develop new insights to higher accuracy and faster speed in learning sparse representations. From an optimization perspective, the sparsity parameter is critical to  proximal operators,
\begin{align}
\prox_H (y)=\arg \min_x \frac{1}{2}||x-y||^2+H(x).
\end{align}
For $\ell_1$ regularization $H(c)=\lambda ||c||_1$, it has a closed form solution $\sign(y)\max (|y|-\lambda,0)$. Although directly enlarging the $\ell_1$ regularizer will force more small weights to be zero within limited time, but this will certainly compromise the accuracy since the larger and effective weights are also penalized strongly. So, we need to find a trade-off between the factors.
There are also some literatures suggesting that the $\ell_1$ norm is in many case inferior to non-convex regularizers,  e.g. smoothly clipped absolute deviation (SCAD)\cite{kim2008smoothly}, the capped $\ell_1$-norm \cite{zhang2010analysis} \cite{zhang2013multi} \cite{gong2012multi} \cite{sun2013robust} which are capable of penalizing part of the parameters lower than a threshold, without penalization on larger weights. They get better performance in applications like learning low rank matrices \cite{zhang2012matrix}\cite{hu2013fast}\cite{han2016multi}\cite{sun2013robust} when applied to suppress small singular values. We continue with a discussion a representative non-convex regularization:
\begin{definition}
The capped $\ell_1$ norm \cite{zhang2010analysis} is defined as
\begin{align}\label{eq:l0}
H(c)=\alpha \sum_{i} \min(|c_i|,\tau), \where \alpha,\tau >0.
\end{align}
\end{definition}
As shown in Figure~\ref{fig:norm}, this formulation  only regularizes those weights that are below a certain threshold $\tau$. For those beyond this value, they can grow arbitrarily without experiencing penalties. It is noteworthy that it is a generalization of $\ell_1$ norm. Particularly, it becomes $\ell_1$ as $\tau \rightarrow \infty$. Generally, with a finite $\tau$, it approximates $\ell_0$ better than $\ell_1$, and therefore leads to higher sparsity in some real-world applications.
The key feature of capped norms is the lack of penalty for weights whose magnitudes are greater than $\tau$, this feature, however, gives a worse performance in the 3D pose recovery with long enough inferences, by the same multi-stage algorithm that we introduced later.
One possible explanation for this question is that an approximated sparse solution, with most elements approaching zero, is perhaps more suitable than a truly sparse one, which has most elements being zero. Since the assumption that any 3D poses are combinations of existing bases is not entirely accurate for all cases. We should also utilize the massive existing works on $\ell_1$ norms that achieve state-of-the-art results with fine-tuned parameters, which is actually useful to combine non-convex regularization with.
With the analysis above in mind, we propose an improved variant, which could keep the advantages of both $\ell_1$ and non-convex norms like capped $\ell_1$.
\begin{proposition}
With $0<\beta < \alpha$, $0<\tau$, the Leaky Capped $\ell_1$ Norm Regularizer (LCNR) is defined as
\begin{eqnarray}\label{eq:norm}
H(c)=\alpha \sum_{i } \min(|c_i|,\tau)+ \beta \sum_{i } \max(| c_i|,\tau).
\end{eqnarray}
\end{proposition}
As shown in Figure~\ref{fig:norm}, \textit{LCNR} is piecewise linear and thus is differentiable (except at a few points). The key difference between the proposed formulation and the standard $\ell_1$ norm and the capped norm is that large weights (\textit{i.e.}~those greater than $\tau$) are still penalized \emph{positively} but
\emph{less heavily}. The parameter selection is simpler - one can first tune $\beta$ for accuracy (or transfer the parameter from tuned $\ell_1$ regularization in existing research) then search $\alpha$ for speed. This is worthwhile since after the procedure is done once, the advantage of faster convergence is enjoyed for thousands of times at deployment. 

\section{ Multi-Stage Optimization }

The aforementioned regularizers, including \textit{LCNR} are non-convex, so some difficulties may arise if we directly apply them in the 3D recovery model. Hence, we propose to use a multi-stage algorithm \cite{zhang2010analysis} to optimize the objective function. At each stage, it constructs a convex upper bound as a surrogate objective. The optimal solution derived in one stage will be used to initialize the optimization at the next stage. Our objective function is same with Eq.(\ref{eq:linearf}). The stage-wise objective is
\begin{eqnarray}\label{eq:unconvex}
F^{l+1}(M,c)=\frac{1}{2}|| Y-  \sum_{i \in \mathcal D} M_i B_i ||_F^2+H^{l+1}(c), 
\end{eqnarray}
in the $(l+1)$-th stage, where $M_i \in \Omega(c_i)$. Here the $H^{l+1}(c)$ is a tight surrogate function, or a tangent at $c^l$ that
\begin{eqnarray}\label{eq:boundfunc}
H(c)\leq H^{l+1}(c), \forall c \in \mathbb R^K, \quad H(c^l)=H^{l+1}(c^l),
\end{eqnarray}
where $c^l$ is the optimal $c$ at the $l$-th stage. If $H(c)$ is set to be LCNR as Eq.(\ref{eq:norm}) or capped $\ell_1$, then
\begin{align*} \label{eq:boundfunc_lcnr}
H^{l+1}(c)=\sum_i |c_i| \cdot \lambda_i^{l},\quad \lambda_i^{l}=\alpha \bi(|c_i^l| \leq \tau)+\beta \bi(|c_i^l| > \tau).
\end{align*}
where $\mathbb I(\cdot)$ is the indicator function that equals to $1$ for truth and $0$ for false. We set the initialization parameter $\lambda_i^0=\beta$, so as to begin with a light regularization for all weights. Although $F(R,c)$ is convex in each stage, the surrogate function is still non-convex by the orthogonal constraint. 
\begin{lemma}
The spectral-norm ball $conv (\Omega (c_i)) = \{ X \in \mathbb R^{2 \times 3} \mid ||X||_2 \leq c_i \}$,
is the tightest convex hull of the Stiefel manifold $\Omega (c_i)$, where $||\cdot ||_2$ represents the spectral norm, which is its largest singular value.  \cite{zhou2016sparse} (\cite{journee2010generalized}, Section 3.4)
\end{lemma}
By relaxing the domain $\Omega$ to $conv(\Omega)$, 
and the recalibration rule of $\lambda_i$ is transferred to  
\begin{eqnarray}
\lambda_i^{l}=\alpha \bi(|| \mh_i^l||_2 \leq \tau)+\beta \bi(||\mh_i^l||_2 > \tau),
\end{eqnarray}
where $\mh_i^l$ is the optimal $M_i$ in the $l$-th stage.
We employ the alternate direction method of multiplier algorithm (ADMM) \cite{boyd2011distributed}\cite{zhou2016sparse}  to attain the stage-wise optimum. The algorithm decomposes an optimization problem with complex regularizations into several smaller problems, each of which keeps an individual copy of global parameters, and all the copies are coupled with the main copy by regularizations. 
We introduce a tensor $V$ as a copy of $M$, 
$U$ as a dual tensor variable and $\mu$ as a stepsize parameter, then rewrite Eq.(\ref{eq:unconvex}) in its augmented Lagrangian formulation
\begin{eqnarray}\label{eq:p2}
F_{\mu}^{l+1}(M,V,U)= \frac{1}{2}||Y -\sum_{i \in \mathcal D} V_i B_i||^2_F  +\sum_{i \in \mathcal D} \lambda_i^l ||M_i||_2 \nonumber \\ + \sum_{i \in \mathcal D} U_i^\top(M_i-V_i) +\frac{\mu}{2} \sum_{i \in \mathcal D} ||M_i-V_i||^2 
\end{eqnarray}
Then the ADMM procedure is applied to solve the subproblem. After the convergence, the multi-stage solver will update the surrogate functions. We denote the inner-iteration superscript as $t$.
Then $M^{t+1}$ is update based on the proximal operator on spectral norms [\cite{proximal}, Section 6.7.2], 
\begin{eqnarray}\label{eq:admm2}
prox_{\lambda} (V_i')= P \diag[\sigma - \lambda_i' \mathcal P_{1} (\sigma/\lambda_i')] Q^\top, 
\end{eqnarray}
where $V_i'=V_i^t -U_i^t/\mu$ and $\lambda_i'=\lambda_i^l/\mu$. Denoting the solution as $prox_{\lambda_i'} (V_i')$,
 $V_i'=P \diag(\sigma) Q^\top$ is the singular value decomposition of $V_i'$, and $\mathcal P_{1}(\cdot)$ is the Euclidean projection onto the $\ell_1$ norm ball.
The update on $V$ and $U$ have closed form solutions,
\begin{align*}
&V^{t+1}_i=(Y B_i^\top+ \mu M_i^{t+1}+ U_i^{t+1})(B_i B_i^\top+ \mu I_3)^{-1} ,\\ &U^{t+1}_i=U_i^t+\mu(M_i^t-V_i^t).
\end{align*}
The convergence property of this algorithm is well studied in \cite{boyd2011distributed}, additionally, we adopt an adaptive policy for stepsize $\mu$ as suggested by [\cite{boyd2011distributed}, Section 3].

\section{Theoretical Analysis}

Generally speaking, the theoretical convergence of non-convex optimization relates the intrinsic data property like restricted isometry property (RIP) or matrix incoherence \cite{buhlmann2011statistics}. 
When combined with non-convex regularizers, even convex problems need specialized analysis, for each problem, 
\textit{e.g.}, multi-task feature learning \cite{gong2012multi}\cite{tang2016capped}, matrix completion \cite{gao2015robust} and \cite{jiang2015robust}\cite{sun2013robust}\cite{han2016multi}\cite{zhang2010analysis}\cite{zhang2013multi}. Our problem is harder since the loss function is also non-convex. 

In this section, we prove that for optimizing the objective function in Eq.\ref{eq:linearf} with \textit{LCNR} as regularization, with a high probability, the estimated affine matrices converge to the ground truth at nearly linear speed against stages. 
We assume that the ground truth of the 2D pose $\bar Y \in \mathbb R^{2 \times p}$ is expressed as a projection of the combined deformation of 3D pose bases, as $\bar Y = \sum_{i \in \mathcal D} \mb_i B_i$, where $\mb_i$ is the ground truth of deformation matrix $M_i$, for $0 \leq i \leq D$. The observation model is $\quad Y=\bar Y+\delta$, where $\delta \in \mathbb R^{2 \times p}$ is a Gaussian noise, i.e. $\delta_{jk} ~ \mathcal N(0, \sigma^2)$. For notational simplicity, we also set $\gamma=\alpha+\beta$. 
\begin{assumption} For any matrix $Z$, $B_i$, and $M_i \in \mathcal R(s)$ in the restricted set $\mathcal R(s)= \{X \in \mathbb R^{n \times m} \mid rank(X) \leq s \}$, that $||Z-M_i B_i||_F \geq \epsilon$, we define the condition constant $\kappa_i$ as
\begin{eqnarray}
\kappa_i = \min_{M_i \in \mathcal R(s)} ||Z -M_i B_i||_F / ||M_i||_{\star} >0.
\end{eqnarray}
\end{assumption} 

\begin{theorem}\label{th1}
Following the common setting of sparse dictionary learning, we assume that each basis $B_i$ are normalized by row $\ell_2$ norm that $\sum_k B_{irk}^2 =\phi$ for all $i \in \mathcal D, 1 \leq r \leq 3$, where $\phi$ is an constant. For the optimal matrix $\mh_i \in \mathbb R^{2 \times 3}$  in any stage, if we set $\alpha,\beta$ as $(\alpha+\beta) \geq \phi \sqrt{3+e}/2$, then  it holds
\begin{eqnarray}
\frac{1}{2 }||\by -\sum_{i \in \mathcal D} \mh_i B_i||_F^2 \leq \frac{1}{2 }||\by -\sum_{i \in \mathcal D} M_i B_i||_F^2 \nonumber \\
+ \sum_{i \in \mathcal D}(4 \gamma+ \lambda_i^{l} )||M_i-\mh_i ||_2 ,
\end{eqnarray}
with the probability of at least $1-2 D\exp(-\frac{1}{2} (e-3 \ln (1+e/3)))$, where $e$ is a positive scalar that $(\alpha+\beta) \geq \phi \sqrt{3+e}/8$.
\end{theorem}
\textbf{Proof.} 
Recalling that $Y=\by+\delta$ and the property of optimal point $\mh_i$, then we have 
\begin{align}
&\frac{1}{2}||\by -\sum_{i \in \mathcal D} \mh_i B_i||_F^2 \leq \frac{1}{2}||\bar Y-\sum_{i \in \mathcal D} M_i B_i||_F^2  \\
&+\sum_{i \in \mathcal D} \lambda_i^l ||M_i-\mh_i ||_2 + \sum_{i \in \mathcal D} \tr[(M_i - \mh_i) B_i \delta^\top] ,\nonumber
\end{align}
where we use the triangular inequality of spectral norm
\begin{eqnarray}
||M_i||_2-||\mh_i ||_2 \leq || M_i-\mh_i ||_2.
\end{eqnarray}
We first establish the upper bound of $\tr[(\mh_i- M_i)B_i \delta^\top]$. We denote a set of random events  $\{ \mathcal A_{ij} \}$ and define a set of random variables $\{ v_{ijr} \}$ as
\begin{eqnarray}
\mathcal A_{ij}= \{ ||B_{i}^\top \delta_j ||_2 \leq \phi \gamma \},\quad v_{ijr}= \frac{1}{\phi}   \sum_{k=1}^p B_{irk} \delta_{jk} ,
\end{eqnarray}
where $B_{irk}$ is the element in the $r$-th row and $k$-th column of $B_i$. Since $B_i$ is normalized, $v_{ijr}$ are i.i.d. Gaussian variables following $\mathcal N(0,1)$. Then 
$\sum_{r=1}^3 v_{ijr}^2$ is a chi-squared random variable with $d=3$ degree of freedom. By choosing $\lambda$ according to Theorem \ref{th1}, we have
\begin{align*}
&Pr(\frac{1}{2} ||B_{i} \delta_j^\top||_2 > \gamma)=Pr(\sum_{r=1}^3 (\sum_{k=1}^p B_{irk} \delta_{jk})^2 >4 \gamma^2 ) \nonumber \\
\leq & Pr(\sum_{r=1}^3 v_{ijr}^2 >3+e) \leq \exp(-\frac{1}{2} \theta(e)) ,
\end{align*}
where $\theta(e) =e- 3 \ln(1+e/3)$ and the second inequality is due to the chi-squared distribution \cite{chen2011integrating}. Denoting $\mathcal A= \bigcap_{i=1}^D \bigcap_{j=1}^2  \mathcal A_{ij} $,  we also denote $\mathcal A^c$ as its complementary set and $|\mathcal A|$ as its cardinality,  then
\begin{eqnarray}
Pr(\mathcal A)= 1-\bigcup_{i=1}^D \bigcup_{j=1}^2  \mathcal A_{ij}^c  
\geq 1-2D \exp(-\frac{1}{2} \theta(e)) .
\end{eqnarray}
Denoting $M_{ir}$ as the $r$-th row of $M_i$, we can derive an upper bound on $\tr[(\mh_i- M_i)B_i \delta^\top]$ under the event $\mathcal A$, 
\begin{align*}
&\tr[(M_i- \mh_i) B_i \delta^\top] =\sum_{r=1}^2 \sum_{j=1}^2 (M_{ir}- \mh_{ir})^\top B_i \delta_j^\top \nonumber \\
 \leq &\sum_{r=1}^2 \sum_{j=1}^2 ||M_{ir}- \mh_{ir}||_2 ||B_i \delta_j^\top||_2  
\leq 4 \gamma ||M_i- \mh_i||_2 ,
\end{align*}
where we apply the Cauchy-Schwarz inequality and the relation between Frobenius norm and spectral norm. By substituting this back 
 we get the proof.

A very important issue with regarding the theoretical analysis is that there has been no shared definition on the recovered error, so it will still be an open and unsolved problem about what is the optimal metric to measure the suboptimality of an estimation. We propose to characterize the spectral norms of the difference matrices between the estimated affine matrices and the corresponding ground-truth. 
Then we proceed to our main theorem.
\begin{definition}
Let $\mh^{l+1}_i$ be the optimal solution at the $(l +1)$-th stage, and $\mh^l_i$ be the one at the $l$-th stage accordingly. We define $W_i^l=\bar M_i - \mh_i^l$, and a function $\mathcal L$ on set $\mathcal S \subseteq \mathcal D$.
We use some constants $\{ \kappa_i \}$ to characterize the condition of dataset, which the convergence speed relies on
\begin{eqnarray}
\kappa \triangleq \min_i \kappa_i,\quad \kappa_i \triangleq \min_{M_i \in \Omega(c)} ||\bar Y -M_i B_i||_F / ||M_i||_{\star} >0, \nonumber
\end{eqnarray}
we also denote two subsets of $\mathcal D$ for convenience,
\begin{eqnarray}
\mathcal E= \{i \in \mathcal D \mid  ||\mb_i||_2 \neq 0\}, \mathcal F= \{i \in \mathcal D \mid ||\mb_i||_2 \leq 2\tau \}.\nonumber
\end{eqnarray}
which charactize the effective components in dictionary. Then an adaptive definition of the estimation error on $\mathcal S$ is
\begin{eqnarray}
\mathcal L_{l}(\mathcal S)=\sqrt{\sum_{i \in \mathcal S} ||\bar M_i - \mh_i^l||_2^2}.
\end{eqnarray} 
\end{definition}

\begin{theorem} \label{th2}
	
If we choose $\alpha,\beta$ as in Theorem \ref{th1} and set $\tau > (\alpha+ \beta) / \kappa^2$, the following inequality stands
\begin{eqnarray}
\mathcal L_{l+1}(\mathcal D) \leq a^l \mathcal L_{0}(\mathcal D) + \frac{b}{1-a}, 
\end{eqnarray}
with a probability of at least $1-2D \exp(-\frac{1}{2} (e-3 \ln (1+e/3)))$,
where $(\alpha+\beta) \geq \phi \sqrt{3+e}/8$, $a= (\alpha+\beta) /(\kappa^2 \tau)$, and $b=\left((\alpha+\beta) \sqrt{D} +\alpha \sqrt{|\mathcal F|}+\beta \sqrt{|\mathcal E|} \right)  /(\kappa^2 \tau)$.
\end{theorem}


\textbf{Proof.} We denote $W^{l}_i=\mh_i^l-\mb_i$ for convenience. We apply Theorem \ref{th1} in stage $(l+1)$ and substitute $M$ by its ground truth $\mb$ , then get
\begin{eqnarray}\label{eq:1}
\frac{1}{2}||\bar Y-\sum_{i \in \mathcal D} \mh^{l+1}_i B_i||_F^2 \leq
\sum_{i \in \mathcal D} ( \gamma+ \lambda_i^l) ||W^{l+1}_i ||_2 ,
\end{eqnarray}
where we use $\bar Y= \sum_{i \in \mathcal D} \mb_i B_i$. We define a set $\mathcal G= \{i \in \mathcal D \mid  ||\mh_i^l||_2 \leq \tau \}$ to separate the weights, and
\begin{eqnarray}
\alpha_i^l=\alpha \bi( i \in \mathcal G),\quad  \beta_i^l=\beta \bi(i \in \mathcal G^c),
\end{eqnarray}
then there is $\lambda_i^l=\alpha_i^l+\beta_i^l$.
Then we establish a bound by Cauchy-Schwarz inequality,
\begin{eqnarray}\label{eq:2}
\sum_{i \in \mathcal D} (\lambda_i^l + \gamma) ||W^{l+1}_i ||_2 =\sum_{i \in \mathcal D} (\alpha_i^l+\beta_i^l+  \gamma) ||W^{l+1}_i ||_2 \nonumber \\
\leq ( \gamma \sqrt{D} +\alpha \sqrt{|\mathcal G|} +\beta \sqrt{|\mathcal G^c|}) \mathcal L_{l+1}(\mathcal D) . 
\end{eqnarray}
By the rule of set operation and the definition of $\mathcal G$ and $\mathcal F$,
\begin{align*}
| \mathcal G |=| \mathcal G \cap \mathcal F|+| \mathcal G \cap \mathcal F^c|,\where  | \mathcal G \cap \mathcal F|
\leq |\mathcal F|,\\
\tau^2| \mathcal G \cap \mathcal F^c| \leq  \sum_{i \in \mathcal G \cap \mathcal F^c} ||\bar M_i-\mh_i^l||_2^2 \leq  \mathcal L_{l}^2(\mathcal G \cap \mathcal F^c) ;
\end{align*}
by the inequality $||\bar M_i-\mh_i^l||_2 \geq ||\bar M_i||_2-||\mh_i^l||_2 \geq \tau$, substituting them back to Eq.(\ref{eq:2}), we get
\begin{eqnarray}
\sum_{i \in \mathcal D} \alpha_i^l ||W_i^{l+1} ||_2 
\leq \alpha \sqrt{|\mathcal F| +  \mathcal L_{l}^2 (\mathcal F^c)/\tau^2} \mathcal L_{l+1} (\mathcal D) \\
\leq (\alpha \sqrt{|\mathcal F|}+ \frac{\alpha }{\tau} \mathcal L_{l}(\mathcal F^c) )  \mathcal L_{l+1} (\mathcal D)  ,
\end{eqnarray}
where in the last inequality we use $\sqrt{a^2+b^2} \leq a+ b$ for $a,b \geq 0$. A similar result holds for another part of Eq.(\ref{eq:2}) as
\begin{eqnarray}
\sum_{i \in \mathcal D} \beta_i^l ||W_i^{l+1}||_2 
\leq (\beta \sqrt{|\mathcal E|}+  \frac{\beta }{\tau} \mathcal L_{l}(\mathcal E^c) )  \mathcal L_{l+1} (\mathcal D) .
\end{eqnarray}
Substituting them back to Eq.(\ref{eq:1}), there is
\begin{align*}
&\frac{1}{2}||\bar Y-\sum_{i \in \mathcal D} \mh^{l+1}_i B_i||_F^2  
\leq \sum_{i \in \mathcal D} ( \gamma +\lambda_i^l) ||W^{l+1}_i ||_2   \nonumber \\
\leq &\left((\alpha+\beta) \sqrt{D} +\alpha \sqrt{|\mathcal F|}+\beta \sqrt{|\mathcal E|} + \frac{1}{\tau} \mathcal L_{l}(\mathcal D) \right)  \mathcal L_{l+1} (\mathcal D) ,
\end{align*}
Recall the definition of $\{\kappa_i\}$ and substitute  $Z=\mb_i B_i$, 
\begin{eqnarray}
\kappa_i^2   ||W_i^{l+1} ||_{2}^2 \leq \kappa_i^2  ||W_i^{l+1}||_{\star}^2 \leq \frac{1}{2}||W_i^{l+1} B_i||_F^2,
\end{eqnarray}
where we use $||X||_F \leq ||X||_2$. Denoting  $\kappa =\min_i \kappa_i$, then
\begin{eqnarray}
2\kappa^2 \sum_{i \in \mathcal D}  ||W_i^{l+1} ||_{2}^2 \leq \sum_{i \in \mathcal D} ||W_i^{l+1} B_i||_F^2
\leq ||\sum_{i \in \mathcal D} W_i^{l+1} B_i||_F^2. \nonumber
\end{eqnarray}
Substituting this to Eq.(\ref{eq:1}) and combining for $i \in \mathcal D$,
\begin{align*}
\kappa^2 \mathcal L_{l+1}^2 (\mathcal D) \leq \left((\alpha+\beta) \sqrt{D} +\alpha \sqrt{|\mathcal F|}+\beta \sqrt{|\mathcal E|} \right)  \mathcal L_{l+1} (\mathcal D) .
\end{align*}
where we apply $\max( |\mathcal E|,|\mathcal F|) \leq D$. Recalling the definition of $a$ and $b$, we obtain
\begin{eqnarray}
\mathcal L_{l+1}(\mathcal D) \leq a  \mathcal L_{l}(\mathcal D) +b \leq a^{l+1} \mathcal L_{0}(\mathcal D) +b \frac{1-a^{l+1}}{1-a} 
\end{eqnarray}
by the pre-setting $0 < a <1$, we obtain the main theorem.
\subsection{Discussion}

Theorem \ref{th2} establishes the convergence property of the estimation error in terms spectral norm, 
since the target is to approximate the ground truth of $\{M_i\}_i$, the algorithm should be less sensitive to the initial values. 

Although Theorems \ref{th1}\&\ref{th2} are actually founded on our specific regularizer and optimizer, we are still able to see the fundamental factors and about how they are going to affect the recovery error. For example, the convergence rate depends on $\mathcal E$ and $\mathcal F$, which are like the \textit{support} of an over-complete dictionary of signals. The regularization, $\alpha,\beta$ have to be in the same order as the norm $\phi$ of pose in dictionary.  
In addition, $\kappa$ is an important constant that relates to the converence speed. 

By analysing the regularization factor, we see that by comparing LCNR of $(\alpha_0,\beta_0,\tau_0)$ with the capped $\ell_1$ norm regularizer that $(\alpha=\alpha_0,\beta=0,\tau_0)$, we see that although the convergence constant $a$ is larger, the probability of the aforementioned inequality holds increases (as $e$ is enlarged); and a differently parameterized LCNR of $(\alpha_0-\delta,\beta_0+\delta,\tau_0)$ may leads the convergence rate to vary due the inherently difference between $|\mathcal E|$ and $|\mathcal F|$. Note that $\ell_1$ is also included in this analysis as $\delta \rightarrow (\alpha_0- \beta_0)/2$. So the flexibility of LCNR improves the potential accuracy by further tuning.

%

\section{Experiments}

\subsection{  Recovery from known 2D poses}

We conducted the experiments to verify the effectiveness of leaky capped $\ell_1$ regularization in learning sparse weights for 3D pose recovery. Our proposed algorithms are compared against state-of-the-art baselines in \cite{zhou2016sparse}, based on the code from its generous authors. We must point it out that, even there exist other methods, like projected matching pursuit \cite{ramakrishna2012reconstructing} and the alternating manifold minimization method,  they were proved inferior to this baseline in extensive comparisons in \cite{zhou2016sparse}, and testing against them show no evidence of improvement of our regularizers since this will make an unfair comparison. 

We use the CMU motion capture dataset \cite{mocap} for both training and testing.
The poses are annotated by 3D locations of 15 landmarks, as $S \in \mathbb R^{3 \times 15}$, and landmarks are at anatomical joints of human, like head,  shoulders, elbows,  hips, ankles and etc.  The dataset contains various kinds of action. As there are large external variations across actions, we take every single action into an analysis, but using the same pose dictionary. We use 300 frames of each action as the test set. The rest of frames are used as training set for building pose dictionary. We set $D=128$ to construct an over-complete dictionary by common sparse coding algorithm with $\ell_1$ regularization, and the training data are pre-aligned by the Procrustes method used in \cite{ramakrishna2012reconstructing}. The 2D poses for test set are synthesized from the ground truth 3D poses at different angles across 360 degrees. The recovery error is measured by the Frobenius norm $||\hat S-S||_F$ from the recovered 3D pose $\hat S$ with the ground truth $S$.

\begin{figure}[!htbp]\label{fig:pose} 
  \vspace{-10pt}
  \begin{center}
    \includegraphics[width=0.5 \textwidth]{
    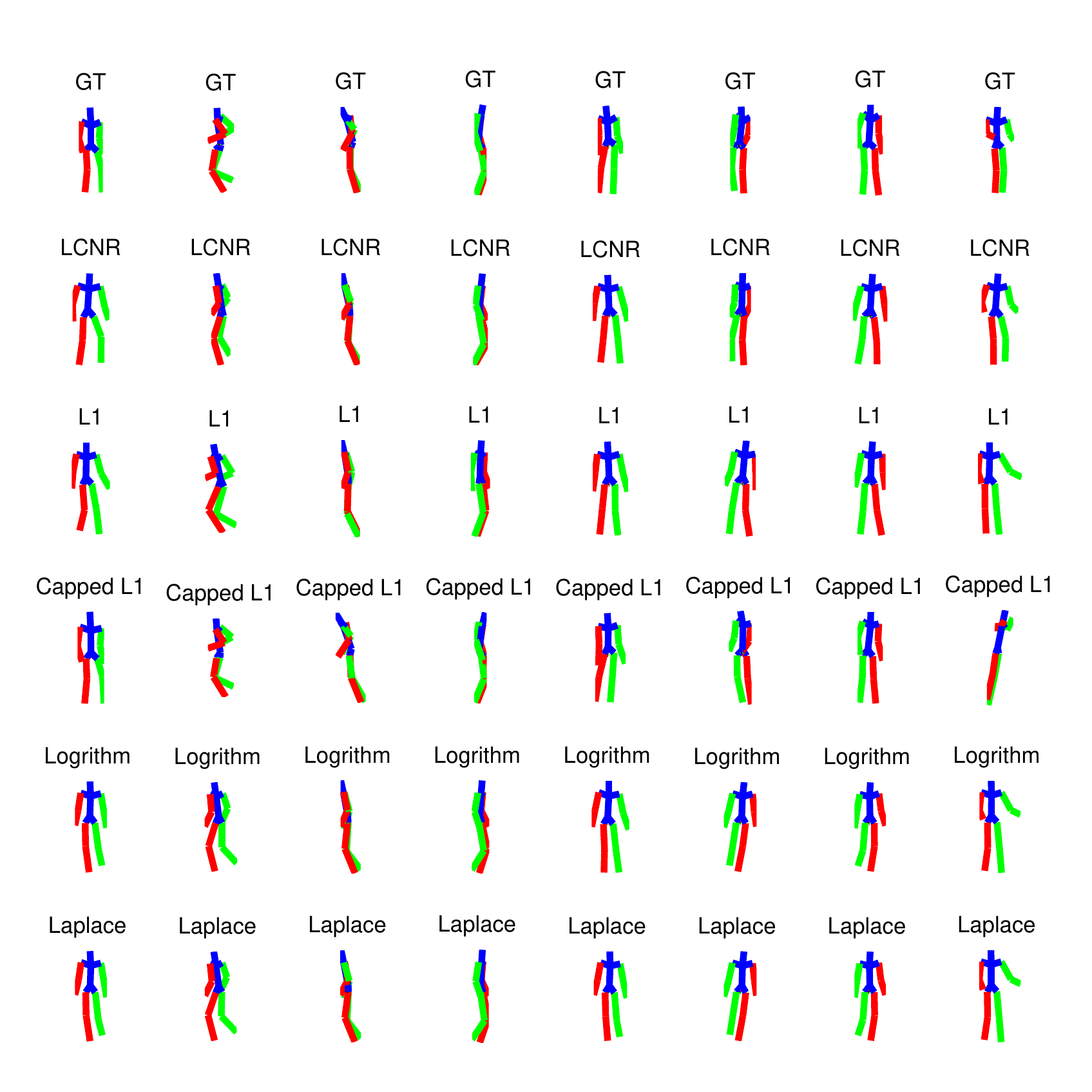}
  \end{center}
  \vspace{-10pt}
  \caption{Recovery results of different kinds of actions, by using LCNR and other (non)convex regularizations, and the ground truth pose (GT) on the top row.}
  \vspace{-10pt}
\end{figure}
We continue to use the multi-stage algorithm to solve the models, and the constraint on dictionary coefficients $c$ is transferred to the spectral norm of 3D poses $\{M_i\}$.
We conducted experiments comparing the LCNR with other representative (non)convex regularizations, including $\ell_1$ norm, Capped $\ell_1$ norm, logarithm norm, and Laplace norm, as
\begin{align*}
&\text{Logarithm}:R(c;\gamma,\lambda)=\frac{\lambda}{\log(\gamma+1)} \log(\gamma |c|+1),\\
&\text{Laplace}: R(c;\lambda,\gamma)=\lambda \left(1-\exp(-\frac{|c|}{\gamma})\right).
\end{align*}
The regularization parameters are grid searched for the best final performance. The computation complexity of calibrating $\tau$ and $\lambda_i^l$ is considerably smaller than the ADMM parts, the average running time of this part is about $14\%$ of the overall time of each iteration. In Figure (2), we plot the recovered poses by LCNR regularized and $\ell_1$ regularized models, and the ground truth shapes, within a maximum inference iterations of $200$. 
To test the performance against large noise, we also add matrix $[\sigma *\textit{mean}( \textit{abs}(S))* \textit{randn}(\textit{size}(S))]$ to each 3D shape $S$ before generating the 2D observation $Y$, and we put the results in appendix due to limited space.
One can see the proposed model reconstruct much more accurate skeletons than state-of-the-art model. 

 \begin{figure}[!htbp]
   \vspace{-10pt}
\subfigure
{
	\includegraphics[width=0.5\textwidth]{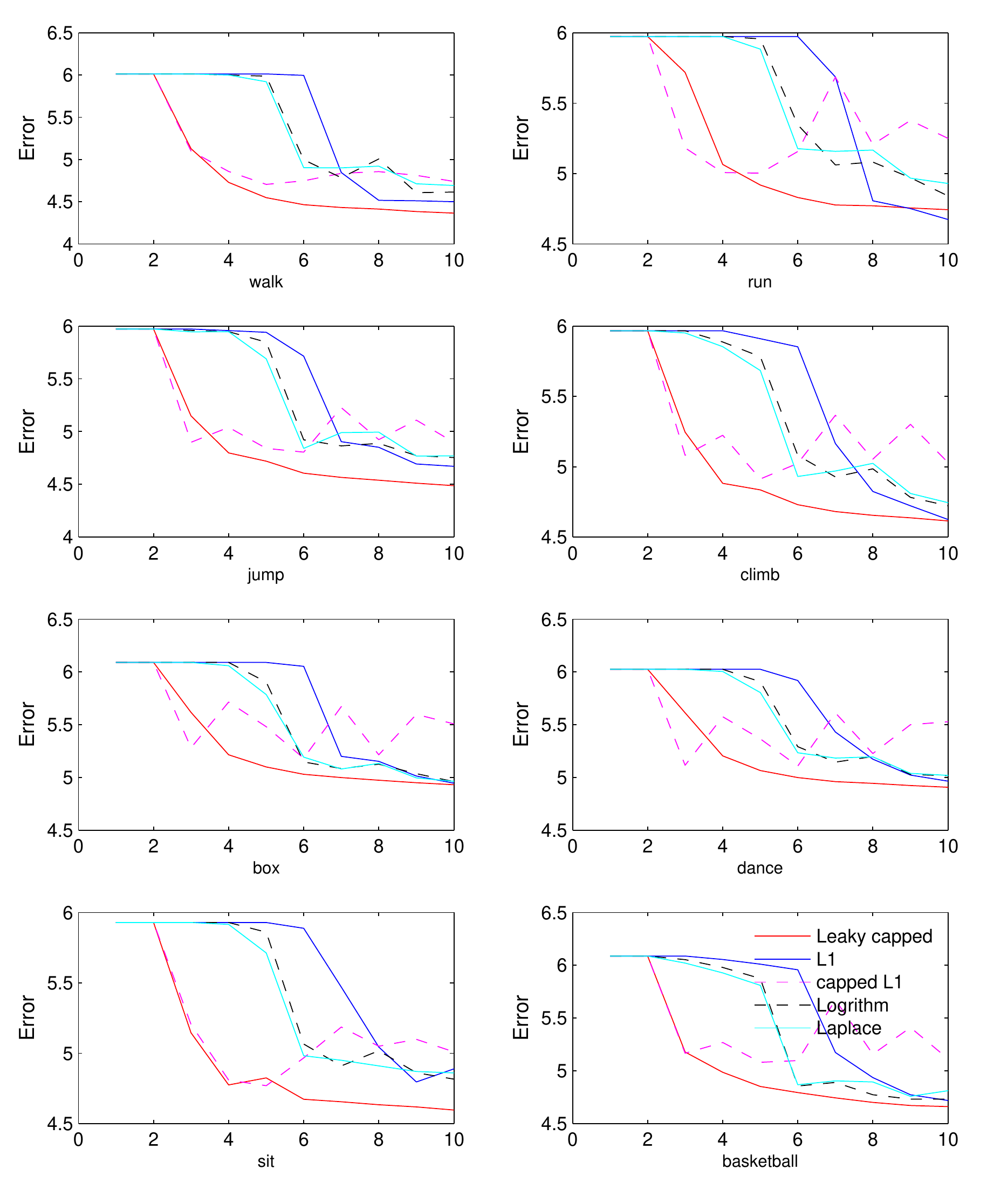}
}
   \vspace{-10pt}
\caption{ Convergence of recovery error with different regularizations. (X-axis: stages (for 10 inner iterations)}
   \vspace{-10pt}
\label{fig:converge}
\end{figure}


\begin{table*}[h]
	\caption{Recovery error on Human36M}
	\label{h36m1}
	\begin{center}
		\begin{tabular}{c | c c c c c c c c c c c }
			\hline
			& Directions &Discussion &Eating & Greeting &Phoning &Photo &Posing &Purchases \\
			\hline
			LinKDE \cite{h36m_pami} &132.7 &183.5 &132.3 &164.3 &162.1 &205.9 &150.6 &171.3\\
			Li et al. \cite{li2015maximum}  &- &136.8 &96.9 &124.7 &- &168.6 &- &- \\
			Tekin et al. \cite{tekin2016direct} & 102.4 &147.7 &88.8 &125.2 &118.0 &182.7 &112.3 &129.1\\
			Du et al. \cite{du2016marker} & 85.0 & 112.6 & 104.9 & 122.0 & 139.0 & 135.9 & 105.9 & 166.1\\
			LCNR &         75.5 &         90.4 &         90.6 &         94.5 &        127.2 &        129.1 &         87.7 &        143.9  \\
			\hline
			& Sitting & SittingDown & Smoking &Waiting &WalkDog & Walking &WalkTogether &Average\\
			\hline
			LinKDE \cite{h36m_pami}  & 151.5  & 243.0  & 162.1  & 170.6  & 177.1  & 96.6  & 127.8  & 162.1\\
			Li et al. \cite{li2015maximum}   & - &  -  & -  & -  & 132.1  & 69.9  & -  & -\\
			Tekin et al. \cite{tekin2016direct}  & 138.8  & 224.9  & 118.4  & 138.7 &  126.2 &  55.0  & 65.7  & 124.9 \\
			Du et al. \cite{du2016marker} & 117.4 &  226.9  & 120.0  & 117.6  & 137.3  & 99.2  & 106.5  & 126.4 \\
			LCNR &        145.6 &        251.6 &        109.5 &        105.0 &        103.1 &         71.0 &         77.3 &        113.6 \\
			\hline
		\end{tabular}
		\vspace{-10pt}
	\end{center}
	%
	
\end{table*}

By comparing the convergence rate in Figure (\ref{fig:converge}), we see that the non-convex regularizations generally converge faster than the $\ell_1$ because they introduce less estimation bias, so a larger coefficient $\lambda$ can be adopted for acceleration. The capped norm induces faster convergence rate within the beginning 3 stages (30 iterations), however, in the following iterations the estimation error oscillates up and down, indicating an unstable behavior, this indicates that a very sparse solution does not benefit the 3D recovery problem. The $\ell_1$ regularization can achieve good estimation at last, but in the beginning, the error decreases very slow, due to the imbalance between accuracy and speed, but it outperforms capped $\ell_1$, indicating that a certain degree of regularization is critical to larger weights.
We can see for most of the actions, LCNR leads to higher recovery accuracy comparing opponents, achieving the same accuracy requirement within much lesser time, this improvement is significant especially to test phase, which is mostly the case since training phase only needs to be accomplished once.

\subsection{  Recovery from detected 2D poses by CNN}
We tested our algorithm by leveraging the deep features extracted from CNN for estimating 3D pose in video sequences. We use the Human36M dataset \cite{h36m_pami}, which contains about videos of 17 kinds of action, from 11 actors, adding to about 3.6 million frames. The dataset was captured by high-speed motion capture system, and has provided 3D landmark locations and joint angles ground-truth. We follow the testing protocol of previous research on this dataset, using 5 subjects out of 11 (S1,S5,S6,S7,S8) for training and using (S9,S11) for testing, and a downsampling of the original videos to 10 fps is done in advance.  

The dictionary of 3D pose is trained for individual actions, and the size if $D=64$ for all actions. We  follow the expectation-maximization framework of \cite{zhou2016sparseness}, on the probabilistic model described in Eq.(\ref{em}),: in each step, the expectation w.r.t. the 2D poses are calculated as $\mathbb E[Y|I,\theta']=\int {\frac{1}{Z}Pr(I|Y)Pr(W|\theta')}$ where $Z$ is the normalizer and $\theta'$ is the current estimate of the parameters, and in the $M$ step, we  optimize an objective function based on the previous estimation $\theta'$ that $Q(\theta|\theta')=\int{Pr(Y|I,\theta')\ln Pr(I,Y|\theta- \mathcal L_t(\theta))dY}$. The parameters are set to be: $\alpha=1.0$, $\beta=0.25$ and $\tau$ is adaptively set to be the $10$-highest spectral norm of $\{M_i\}_i^D$. To demonstrate the inference speed of our approach, we set the maximum iterations to $300$ and each stage consists of $20$ iterations and one time of recalibration of $\lambda$.

We use the stacked hourglass network \cite{newell2016stacked} for generating the heatmaps. This network takes each frame as input, cascades several blocks of consecutive downsampling, upsampling layers and skip-connect layers, and generates the heatmaps of size $64 \times 64$ for all landmarks using a single output tensor, each channel of which represents a corresponding landmark and shares convolutional features in the previous layers together. The loss function is the $\ell_2$ normalized distance of all landmarks. The predicted 2D landmarks are locations that have the maximum response in the heatmap. The evaluation metric on the testing data is the mean square per landmark error as the average of Euclidean space distance between the estimated landmarks and ground truth. Plus, the translation factor is neutralized by appropriately aligning the root locations of human poses.

We compare recent approaches that also use deep networks for detecting human poses \cite{h36m_pami} \cite{li2015maximum}  \cite{tekin2016direct} \cite{du2016marker}. To support our claim on accelerating inference to real-time applications, we restricted the number of maximum iterations to $300$, which is much smaller than previous settings, e.g. \cite{zhou2016sparseness} has used $1000$ iterations to perform inference, and each iteration costs about the same time with ours. The recovery error in terms of Euclidean distance of all landmarks from different methods is summarized in Table (\ref{h36m1}), where our algorithm is shortened as LCNR. From the table, we see that our algorithm surpasses most of the existing methods. Our empirical results can be improved if combined with 3D end-to-end methods by enforcing consistence of two results  from two approached, 
and have demonstrate a clear advantage under limited time, which is a real demand when deployed to embedded devices with weaker computation power and needs to be real time. Plus, the 3D dictionary based model gives better physics interpretation on the results, not tending to overfit.

\section{Conclusion}

We present a monocular 3D pose recovery with non-convex regularization and a multi-stage optimizer. We obtain an improvement in accuracy and acceleration upon state-of-the-art algorithms. We give a theoretical analysis, making an important step to understanding how those fundamental factors (optimization iterations, regularizations, dictionary, observation noise, ground-truth) are affecting the recovery results of dictionary-based methods, and moving towards the optimal speed-accuracy trade-off.
%
%

\newpage

\bibliographystyle{ieee}
\bibliography{egbib}

\end{document}